\documentclass{article}
\usepackage{spconf,amsmath,graphicx,amssymb}
\usepackage{multirow}
\usepackage{booktabs}

\usepackage{xcolor}

\title{Few-shot Image Classification with Multi-Facet Prototypes}
\name{\begin{tabular}{c}Kun Yan$^{1}$, Zied Bouraoui$^{2}$, Ping Wang$^{1,3,4,*}$, Shoaib Jameel$^{5}$, Steven Schockaert$^{6}$ \end{tabular}\thanks{$^*$ Corresponding author.}}
\address{
    $^{1}$School of Software and Microelectronics, Peking University, China\\
    $^{2}$ CRIL - CNRS \& University Artois, France\\
    $^{3}$National Engineering Research Center for Software Engineering, Peking University, China\\
    $^{4}$Key Laboratory of High Confidence Software Technologies~(PKU), Ministry of Education, China\\
    $^{5}$School of Computer Science and Electronic Engineering, University of Essex, UK\\
    $^{6}$School of Computer Science and Informatics, Cardiff University, UK
}
\begin{document}
%
\maketitle
\begin{abstract}
The aim of few-shot learning (FSL) is to learn how to recognize image categories from a small number of training examples. A central challenge is that the available training examples are normally insufficient to determine which visual features are most characteristic of the considered categories. To address this challenge, we organise these visual features into facets, which intuitively group features of the same kind (e.g.\ features that are relevant to shape, color, or texture). This is motivated from the assumption that (i) the importance of each facet differs from category to category and (ii) it is possible to predict facet importance from a pre-trained embedding of the category names.
In particular, we propose an adaptive similarity measure, relying on predicted facet importance weights for a given set of categories. This measure can be used in combination with a wide array of existing metric-based methods.
Experiments on miniImageNet and CUB show that our approach improves the state-of-the-art in metric-based FSL.
\end{abstract}
\section{Introduction}
\label{sec:intro}
Few-shot learning aims to recognize new image categories given only a few labeled examples. Most existing methods either fall into the meta-learning based~\cite{optimization-as-model,MAML,Meta-SGD} or the metric-based
~\cite{relationnet,protonet,feat,match-net} paradigms. 
In this paper, we focus in particular on metric-based few-shot learning methods, which have received a lot of attention due to their simplicity, flexibility and effectiveness. These methods aim to learn an embedding space in which the similarity of images can be evaluated using a pre-defined metric, such as Euclidean distance. 

Early work on one-shot learning used Siamese networks \cite{siamese} to learn appropriate 
embedding spaces. Beyond one-shot learning, \cite{match-net} introduced an episode-based training strategy and proposed Matching Network, an FSL model which combines a weighted nearest-neighbor classifier with an attention mechanism. ProtoNet~\cite{protonet} first constructs a prototype for each category, by taking the mean of all the corresponding training examples, and then classifies test images based on the nearest prototype. Instead of fixing a pre-defined metric, Relation Network~\cite{relationnet} learns a deep distance metric to compare each query-support image pair. The recently introduced FEAT model~\cite{feat} applies a transformer~\cite{DBLP:conf/nips/VaswaniSPUJGKP17} to contextualize the image features relative to the support set in a given task. 

The aforementioned methods only rely on visual features. 
However, in addition to the training examples, we usually also have access to the names of the image categories to be learned. AM3~\cite{AM3} uses a pre-trained embedding of these category names, using the GloVe word embedding model \cite{DBLP:conf/emnlp/PenningtonSM14}, to predict visual prototypes. TRAML~\cite{traml} also uses GloVe vectors, but instead uses them to get prior information about which categories may be most difficult to separate. More precisely, TRAML uses the word vectors to implement an adaptive margin-based model.

In this paper, we propose a third way in which word vectors may be used to improve metric based FSL models. Our starting point is that we can intuitively think of the visual features as capturing different facets, such as shape, texture, color, etc. The relative importance of these facets moreover strongly depends on the considered image categories.
For example, color-related facets are likely to be important to classify image from the `desert' category, whereas shape-related facets may be more important for images from the `cat' category. Therefore, we propose a method to predict the importance of the different facets for a given category, based on a pre-trained embedding of the category name. 
%
In addition to using GloVe vectors for representing the category names, we also experiment with embeddings that are obtained using the BERT language model \cite{DBLP:conf/naacl/DevlinCLT19}. 
The facets themselves are generated by clustering the coordinates of the visual feature vectors, using a similarity metric that is based on which visual classes can be discriminated well by the different coordinates.


\section{Background on FSL}
\label{ssec:p_s}

We briefly recall the few-shot learning~(FSL) setting. Let a set of base classes $\mathcal{C}_{\textit{base}}$ and a disjoint set of novel classes $\mathcal{C}_{\textit{novel}}$ be given. These two sets differ in the number of available training images: whereas a sufficiently large number of training examples is given for the classes from  $\mathcal{C}_{\textit{base}}$, only a few training examples are available for the classes in $\mathcal{C}_{\textit{novel}}$. The goal of FSL is to obtain a classifier that performs well for the novel classes in $\mathcal{C}_{\textit{novel}}$. Usually, an $N$-way $K$-shot setting is assumed, where there are $N$ novel classes and $K$ training examples per class, where typically $K\in \{1,5\}$.
Rather than using a fixed training-test split, FSL models are typically evaluated using so-called episodes. Specifically, in each test episode, $N$ classes from $\mathcal{C}_{\textit{novel}}$ are sampled, and $K$ labelled examples from each class are made available for training. The remaining images from the $N$ sampled classes are then used as test examples. The support set of an episode is the set of sampled training examples. We write it as $\mathcal{S} = \left\{ (x_{i}^{s}, y_{i}^{s}) \right\}_{i=1}^{n_{s}}$, where $n_{s} = N \times K$, $x_i^s$ are the sampled training examples and $y_i^s$ are the corresponding class labels. Similarly, the \emph{query set} contains the sampled test examples and is written as $\mathcal{Q} = \left\{ (x_{i}^{q}, y_{i}^{q}) \right\}_{i=1}^{n_{q}}$. 
To train FSL models, we also adopt an episode-based approach, as proposed by~\cite{match-net}. The model is thus trained by repeatedly sampling $N$-way $K$-shot episodes from $\mathcal{C}_{\textit{base}}$, rather than by using $\mathcal{C}_{\textit{base}}$ directly. 

In our proposed method, we will rely on ProtoNet~\cite{protonet}, which is one of the most popular metric-based FSL methods. Throughout the paper, we assume that the visual features $f_{\theta}(x)\in \mathbb{R}^{n_{v}}$ of an image $x$ are extracted by a CNN model such as ResNet~\cite{residual}. Following ProtoNet, we construct a visual prototype $\mathbf{v}_{c}$ of a class $c$ by averaging the visual features of all training images from this class, in a given episode $p$:
\begin{equation}
\label{eq1}
	\mathbf{v}_{c} = \frac{1}{K} \sum\{f_{\theta}(x_i^s) \,|\, (x_i^s,c)\in \mathcal{S}_p\}
\end{equation}
where $\mathcal{S}_p = \left\{ (x_{i}^{s}, y_{i}^{s}) \right\}_{i=1}^{n_{s}}$ is the support set of episode $p$.

\section{Method}
\label{sec:method}
In Section \ref{ssec:c_n_e}, we first explain how the BERT language model can be used to obtain pre-trained embeddings of visual classes, based on their names. Subsequently, in Section \ref{secFacetDetection} we explain the visual features can be decomposed into meaningful facets, without introducing additional parameters into the overall model, which is an important requirement in the context of few-shot learning. Finally, Section \ref{secSimilarityComputation} introduces our facet-weighted similarity computation, and explains how this adaptive similarity measure can be used in combination with metric-based models such as ProtoNet.

\subsection{Class Name Embeddings}
\label{ssec:c_n_e}
We now explain how we use BERT \cite{DBLP:conf/naacl/DevlinCLT19} to get vector representations of class names. BERT represents sentences as sequences of so-called word-pieces. These correspond to words and sub-word tokens, where the latter are used to encode rare words. Each word-piece is represented by a static input vector. BERT maps a sequence of input vectors (corresponding to the word-pieces from a given sentence) onto a sequence of output vectors, which intuitively represent the meaning of the word-pieces within the context of the sentence. When representing the input, we can also replace words or phrases by a special [MASK] symbol. The corresponding output vector then captures what the sentence reveals about the missing word or phrase. We follow this approach to obtain class name embeddings.
Specifically, let $\mathcal{C}$ be the set of classes for which we want to learn an embedding. For each class $c\in \mathcal{C}$, we collect a number of sentences $S(c)={s_1,...,s_m}$ that mention the name of this class. In particular, for our experiments we sampled $m=1000$ such sentences from the May 2016 English Wikipedia dump. In each of the sampled sentences, we replace the name of the class by [MASK] and compute the corresponding output vector. The final embedding of the class name is obtained by averaging the resulting $m$ output vectors. In practice, the classes often correspond to WordNet synsets. For each class, we may then have several synonymous names. In such cases, we first get a vector for each name from the synset, in the aforementioned way, and then average the resulting vectors. 


\subsection{Facet Identification}\label{secFacetDetection}
Our aim is to group the coordinates of the visual feature vectors $f_{\theta}(x)$, such that coordinates from the same group intuitively refer to similar aspects.
Formally, we want to find a partition $X_1,...,X_F$ of the set of coordinate indices $\{1,...,n_v\}$. We write $f^i_{\theta}(x)$ for the restriction of $f_{\theta}(x)$ to the indices in $X_i$. For each image $x$ we thus have $F$ different vectors, $f^1_{\theta}(X),...,f^F_{\theta}(X)$, each of which intuitively focuses on a different facet. A key problem is that we want to identify these facets, i.e.\ the partition classes $X_i$, without introducing (too many) additional parameters, to prevent overfitting.
For this reason, we will treat the problem of finding the facets $X_1,...,X_F$ as a clustering problem. To this end, we need a similarity measure between coordinates, capturing how closely related the corresponding visual feature are.

Before explaining how we measure the similarity between different coordinates, we first introduce a measure of coordinate (or feature) importance. In particular, for a given training episode $p$, we determine how important coordinate $i$ is in differentiating the  classes that were sampled.
After computing the prototypes for each class, according to Eq.~\ref{eq1}, we take into account how often the $i^{\textit{th}}$ coordinate of a training example is closer to the $i^{\textit{th}}$ coordinate of the prototype of its own class than to the $i^{\textit{th}}$ coordinate of the prototypes of the other $N-1$ classes. For a vector $\mathbf{a}$, we write $\mathbf{a}[i]$ to denote its $i^{\textit{th}}$ coordinate. The importance of the $i^{\textit{th}}$ coordinate, for the class $c$, is then computed as follows:
\begin{align*}
	a_c^i {=} \sum_{(x_j,c)}\textsc{ReLU}\left(\frac{\sum_{d\in \mathcal{C}_p\setminus\{c\}}\|f_{\theta}(x_j)[i]- \mathbf{v}_{d}[i]\|_2^2}{(N-1)\,\|f_{\theta}(x_j)[i] - \mathbf{v}_{c}[i]\|_2^2} - 1\right)
\end{align*}
where $\mathcal{C}_p$ is the set of classes from the current episode $p$ and $|\mathcal{C}_p|=N$. 
The outer summation is taken over all the available labelled examples $(x_j,c)$ of class $c$ (i.e.\ covering both the support and query set).
Note that the argument of \textsc{ReLU} is positive if $f_{\theta}(x_j)[i]$ is closer to the $i^{\textit{th}}$ coordinate of the prototype of $c$ is than to the $i^{\textit{th}}$ coordinate of the prototypes of the other classes, on average. We then define $a^i = \frac{1}{N}\sum_{c\in\mathcal{C}_p}a^i_c$. In other words, the importance $a^i$ of coordinate $i$, for a given episode, is defined as the average of its importance for the different classes. We repeat this computation for $m$ different episodes, where each time $N$ classes are sampled from $\mathcal{C}_{\textit{base}}$. We then construct an $m \times n_v$ matrix $\mathbf{A}$, where the element on the $j^{\textit{th}}$ row and $i^{\textit{th}}$ column is the importance score $a^i$ that was found for the $j^{\textit{th}}$ episode.


To determine how closely (the visual features corresponding to) two coordinates $i$ and $j$ are related, we measure how strongly their importance scores $a^i$ and $a^j$ are correlated. To this end, we compute the Kendall $\tau$ statistic between the $i^{\textit{th}}$ and $j^{\textit{th}}$ column of $\mathbf{A}$. Let us write $e_{ij}\in [-1,1]$ for the resulting value. Note that $e_{ij}$ is close to 1 if the coordinates $i$ and $j$ are important for the same episodes, whereas a value close to -1 would mean that whenever $i$ is important, $j$ tends to be unimportant, and vice versa. 
Then, we use average-link agglomerative hierarchical clustering to partition the set $\{1,...,n_v\}$ into the facets $X_1,...,X_F$, where the values $e(i,j)$ are used to measure similarity. The number of facets $F$ is treated as a hyper-parameter. 

\subsection{Similarity Computation}\label{secSimilarityComputation}
When computing the similarity between a query image and a class prototype, in a metric-based FSL model such as ProtoNet, we now want to take into account the relative importance of different facets. In particular, for a given test episode, we first predict the importance of each facet using the embeddings $\mathbf{n}^c$ that we obtained using BERT, as explained in Section \ref{ssec:c_n_e}. 
To this end, we introduce a facet-importance generation network $g_e$, which maps $\mathbf{n}^c$ onto an $F$-dimensional vector, intuitively capturing the importance of each of the $F$ facets for class $c$:
 \begin{align}
     \mathbf{\mathbf{b}_c} = g_e(\mathbf{n}^c)
 \end{align} 
 where $\mathbf{\mathbf{b}_c} \in \mathbb{R}^{F}$. We obtain the final facet importance weights by applying a softmax layer to $\mathbf{b_c}$  as follows:
 \begin{align}
	(\eta_c^1,...,\eta_c^F) = \textsc{softmax}(\mathbf{b_c})
\end{align}
Finally, we write $\eta^i$ for the average of the facet-importance scores across the set of classes $\mathcal{C}_p$ from the given test episode, i.e. 
$$
\eta^i = \frac{1}{N} \sum_{c\in\mathcal{C}_p} \eta_c^i
$$
Given these importance scores, we can measure the distance between a query image $q$ and the prototype of class $c$ as a weighted sum of facet-specific distances, as follows: 
\begin{align}\label{eqFacetDistance}
	\textit{fdist}(q,c) = \sum_{i=1}^{F} \eta^i\, \|f_{\theta}^i(q) - \mathbf{v}^i_c\|_2^2
\end{align}
Rather than using $\textit{fdist}(q,c)$ directly, we combine $\textit{fdist}(q,c)$ with the standard Euclidean distance, as used in ProtoNet, as follows:
\begin{align}
	\textit{dist}(q,c) =  \| f_{\theta}(q) - \mathbf{v}_c\|_2^2 + \lambda\, \textit{fdist}(q,c)
\end{align}
where $\lambda\geq 0$ is a hyper-parameter to control the contribution of the facet-weighted similarity computation. Note in particular that for $\lambda=0$ we recover the original ProtoNet model. In the same way, we can also combine \textit{fdist}(q,c) with the distance metrics from other metric-based FSL models, such as FEAT.

\section{Experiments}
\label{experiment}
In this section, we evaluate whether standard FSL methods can be improved by incorporating our proposed facet-specific distance measure.

\subsection{Dataset}
We conduct experiments on two benchmark datasets: miniImageNet~\cite{match-net} and CUB~\cite{CUB}. The miniImageNet dataset is a subset of ImageNet~\cite{ImageNet}. It consists of 100 classes. For each of these classes, 600 labeled images of size 84 $\times$ 84 are provided. We adopt the common setup introduced by~\cite{optimization-as-model}, which defines a split of 64, 16 and 20 classes for training, validation and testing respectively. The CUB dataset contains 200 classes and 11\,788 images in total. We used the splits from \cite{closer-look}, where 100 classes are used for training, 50 for validation, and 50 for testing.

\subsection{Training details}
We evaluate our method on 5-way 1-shot and 5-way 5-shot settings. Following the standard training strategy, we train 60\,000 episodes in total for miniImageNet and 40\,000 episodes for CUB. 
We use $m=5000$ episodes to compute the Kendall $\tau$ statistics.
During the test phase, 600 test episodes are generated. We report the average accuracy as well as the corresponding 95\% confidence interval over these 600 episodes. 
To obtain the BERT-based vectors, we use the BERT-large-uncased model, which yields 1024 dimensional vectors. As the backbone network for producing the visual feature embeddings, we consider \emph{ResNet-10}~\cite{residual} for ablation study,  \emph{ResNet-12} and Conv-64~\cite{protonet} for miniImageNet and CUB respectively for fair comparison with other methods. The remaining parameters were selected based on the validation set. This resulted in a choice of $\lambda=10$ for miniImageNet and $\lambda=8$ for CUB, while $F$ was set to 7 for miniImageNet and 5 for CUB.

\subsection{Ablation Study}
Our main hypothesis in this paper is that meaningful facet-importance scores can be predicted from class name embeddings. In Table \ref{table4} we directly test this hypothesis by comparing (i) the standard ProtoNet model, (ii) a variant of our model in which GloVe vectors are used instead and (iii) the proposed model based on BERT. As can be seen, both variants of the facet-guided method outperform the ProtoNet baseline, with the BERT-based vectors outperforming GloVe vectors.

\begin{table}[t]
\footnotesize
\begin{center}
\caption{The 5-way 5-shot accuracies~(\%) with a 95\% confidence interval on the miniImageNet dataset.}
\label{table4}
\begin{tabular}{llcc}
\toprule
\textbf{Method} & \textbf{Backbone}
& \textbf{Word Embeddings} & \textbf{5-way 5-shot}\\
\midrule
ProtoNet            & ResNet-10 & None  & 73.24 $\pm$ 0.63\\
Ours(ProtoNet)      & ResNet-10 & GloVe & 74.10 $\pm$ 0.61\\
Ours(ProteNet)      & ResNet-10 & BERT  & 75.24 $\pm$ 0.76\\

\bottomrule
\end{tabular}
\end{center}
\vspace{-3mm}
\end{table}

\begin{table}[t]
\footnotesize
\begin{center}
\caption{The mean accuracies~(\%) with a 95\% confidence interval on the miniImageNet dataset.}
\label{table2}
\begin{tabular}{llcc}
\toprule
\textbf{Method}  & \textbf{Backbone} &\textbf{5-way 1-shot} &\textbf{5-way 5-shot}\\
\midrule
MAML~\cite{MAML}   	        & Conv-64     & 48.70 $\pm$ 1.75 & 63.15 $\pm$ 0.91\\
Reptile~\cite{Reptile}  	    & Conv-64     & 47.07 $\pm$ 0.26 & 62.74 $\pm$ 0.37\\
LEO~\cite{meta-latent-embedding} & WRN-28 & 61.76 $\pm$ 0.08 & 77.59 $\pm$ 0.12\\
MTL~\cite{mtl}  	        & ResNet-12     & 61.20 $\pm$ 1.80 & 75.50 $\pm$ 0.80\\
MetaOptNet-SVM~\cite{mlwd}  & ResNet-12  & 62.64 $\pm$ 0.61 & 78.63 $\pm$ 0.46\\

\midrule
Matching Net~\cite{match-net} & Conv-64  & 43.56 $\pm$ 0.84 & 55.31 $\pm$ 0.73\\
ProtoNet~\cite{protonet} & Conv-64  & 49.42 $\pm$ 0.78 & 68.20 $\pm$ 0.66\\
RelationNet~\cite{relationnet} & Conv-64  & 50.44 $\pm$ 0.82 & 65.32 $\pm$ 0.70\\
ProtoNet~\cite{protonet}  & ResNet-12  & 56.52 $\pm$ 0.45 & 74.28 $\pm$ 0.20\\
TADAM~\cite{tadam}        & ResNet-12   & 58.50 $\pm$ 0.30 & 76.70 $\pm$ 0.38\\

AM3(ProtoNet, BERT)       & ResNet-12   & 62.11 $\pm$ 0.39 & 74.72 $\pm$ 0.64\\
AM3(ProtoNet, {GloVe})    & ResNet-12   & 62.43 $\pm$ 0.80 & 74.87 $\pm$ 0.65\\
AM3(ProtoNet++)~\cite{AM3}  & ResNet-12 & 65.21 $\pm$ 0.49 & 75.20 $\pm$ 0.36\\
TRAML(ProtoNet)~\cite{traml} & ResNet-12 & 60.31 $\pm$ 0.48 & 77.94 $\pm$ 0.57\\
DSN-MR~\cite{DSN-MR}     & ResNet-12 & 64.60 $\pm$ 0.48 & 79.51 $\pm$ 0.50\\
DeepEMD~\cite{deepemd}   & ResNet-12 & 65.91 $\pm$ 0.82 & 82.41 $\pm$ 0.56\\
FEAT~\cite{feat}         & ResNet-12 & 66.78 & 82.05 \\

\midrule
Ours(ProtoNet) & ResNet-12 &  63.21 $\pm$ 0.37 &  77.84 $\pm$ 0.64\\
Ours(FEAT)  & ResNet-12   &  \textbf{67.24 $\pm$ 0.58} & \textbf{82.51 $\pm$ 0.66}\\							
\bottomrule
\end{tabular}
\end{center}
\vspace{-3mm}
\end{table}

\begin{table}[t]
\footnotesize
\begin{center}
\caption{The mean accuracies~(\%) with a 95\% confidence interval on the CUB dataset.}
\label{table3}
\begin{tabular}{llcc}
\toprule
\textbf{Method} & \textbf{Backbone}
& \textbf{5-way 1-shot} & \textbf{5-way 5-shot}\\
\midrule
MAML          & Conv-64 & 55.92 $\pm$ 0.95 & 72.09 $\pm$ 0.76\\
Matching Net  & Conv-64 & 61.16 $\pm$ 0.89 & 72.86 $\pm$ 0.70\\
ProtoNet      & Conv-64 & 51.31 $\pm$ 0.91 & 70.77 $\pm$ 0.69\\
RelationNet   & Conv-64 & 62.45 $\pm$ 0.98 & 76.11 $\pm$ 0.69\\
Baseline++    & Conv-64 & 60.53 $\pm$ 0.83 & 79.34 $\pm$ 0.61\\
SAML~\cite{saml}   & Conv-64 & 69.35 $\pm$ 0.22 & 81.37 $\pm$ 0.15\\
DN4~\cite{revisit} & Conv-64 & 53.15 $\pm$ 0.84 & 81.90 $\pm$ 0.60\\
\midrule
Ours(ProtoNet)  & Conv-64  & {\bf 69.52 $\pm$ 0.76}  & {\bf 82.34 $\pm$ 0.66}\\
\bottomrule
\end{tabular}
\end{center}
\vspace{-2mm}
\end{table}

\subsection{Comparison with State-of-the-art}
Tables \ref{table2} and \ref{table3} compare our method against state-of-the-art methods on miniImageNet and CUB respectively. As can be seen in Table \ref{table2}, when using ProtoNet as the base model, our method significantly outperforms the standard ProtoNet model, both in the 1-shot and 5-shot settings, with the difference being most pronounced in the 1-shot case. When FEAT is used as the base model, we again see a consistent improvement compared to the standard FEAT model. This version of our model also achieves the best results overall. The AM3 and TRAML models are of particular interest, because they also incorporate word vectors. For the case of AM3, in addition to the standard variant, which uses GloVe, we also obtained results with our BERT based class embeddings. This shows that the improvements we obtain over AM3 are not only due to the change from GloVe to BERT. In fact, in the case of AM3, the BERT based vectors actually underperform the GloVe vectors, presumably as as result of their much higher dimensionality. For TRAML, we only report the published results, as we did not have access to the implementation of this model. For the results on CUB in Table \ref{table3}, we can see that our method improves on the standard ProtoNet model in a very substantial way. Furthermore, our model again achieves the best results overall.



\section{Conclusion}
We have proposed a method to improve the performance of metric-based approaches for few-shot image classification by taking embeddings of class names into account. 
Different from existing methods, we use these class name embeddings to predict the performance of different facets, and then measure the distance between images and prototypes as a weighted sum of facet-specific distances. The resulting facet-based distance can then be combined with a standard distance, e.g.\ the Euclidean distance in the case of ProtoNet.
Experiments on two standard datasets showed consistent improvements compared to state-of-the-art methods. We also found that class name embeddings obtained from the BERT language model yielded better results than GloVe vectors, despite the ongoing popularity of the latter in FSL models.

\vfill\pagebreak

\bibliographystyle{IEEEbib}
\bibliography{few-shot-learning}

\end{document}